\documentclass{article}

\usepackage[final]{nips_2017}
\usepackage{latexsym}
\usepackage{algorithm, algorithmcx, caption}
\usepackage{graphicx}
\usepackage{ulem}
\usepackage{framed}
\usepackage[justification=centering]{caption}

\usepackage[utf8]{inputenc} 
\usepackage[T1]{fontenc}    
\usepackage{hyperref}       
\usepackage{url}            
\usepackage{booktabs}       
\usepackage{amsfonts}       
\usepackage{nicefrac}       
\usepackage{microtype}      

\title{Simple and Effective Dimensionality Reduction for Word Embeddings}

\author{
  Vikas Raunak\\
  Microsoft India, Hyderabad\\
  \texttt{viraun@microsoft.com} \\
}

\begin{document}

\maketitle

\begin{abstract}
Word embeddings have become the basic building blocks for several natural language processing and information retrieval tasks. Pre-trained word embeddings are used in several downstream applications as well as for constructing representations for sentences, paragraphs and documents. Recently, there has been an emphasis on further improving the pre-trained word vectors through post-processing algorithms. One such area of improvement is the dimensionality reduction of the word embeddings. Reducing the size of word embeddings through dimensionality reduction can improve their utility in memory constrained devices, benefiting several real-world applications. In this work, we present a novel algorithm that effectively combines PCA based dimensionality reduction with a recently proposed post-processing algorithm, to construct word embeddings of lower dimensions. Empirical evaluations on 12 standard word similarity benchmarks show that our algorithm reduces the embedding dimensionality by 50\%, while achieving similar or (more often) better performance than the higher dimension embeddings.   
\end{abstract}

\section{Introduction}

Word embeddings are distributed and dense real-valued representations of words as low dimensional vectors, that geometrically capture the semantic ``meaning" of a word, along with several linguistic regularities such as analogy relationships. Such embeddings (e.g. Glove [12], word2vec Skip-Gram [7]) are learned from unlabeled text corpora and have found great use in several natural language processing and information retrieval tasks [9]. Given their widespread utility, recently there has been an emphasis on applying post-processing algorithms on the pre-trained word vectors to further improve their quality. For example, algorithm in [8] tries to inject antonymy and synonymy constraints into vector representations, while [4] tries to refine word vectors by using relational information from semantic lexicons such as WordNet. [3] tries to remove the biases (e.g. gender biases) present in word embeddings and [10] tries to ``denoise" word embeddings by strengthening salient information and weakening noise. In particular, the post-processing algorithm in [9] tries to improve word embeddings by projecting the embeddings away from the most dominant directions and considerably improves the embeddings' performance by making them more discriminative.

Another issue related with word embeddings is their size [6]. For example, loading a word embedding matrix of 2.5M tokens takes up to 6 GB memory (for 300-dimensional vectors, on a 64-bit system). Such large memory requirements impose significant constraints on the practical use of word embeddings, especially on mobile devices where the available memory is often highly restricted. [6] tries to ameliorate this situation by using limited precision representation during word embedding use and training while [1] tries to compress word embeddings using different compression algorithms. Our approach differs from both these works as we directly try to reduce the dimensionality of word embeddings rather than using limited precision representation or compressing individual vector values. In the next section, we explain our algorithm and describe with an example the choices behind its design. The evaluation results are presented subsequently. 

\section{The Algorithm}
In this section, first, the post-processing algorithm from [9] is explained in section 2.1. Our algorithm, along with its motivations is explained in section 2.2. Section 2.3 presents the experimental results on pre-trained Glove \footnote{Available at https://github.com/facebookresearch/fastText/.} and fastText Skip-Gram embeddings \footnote{Available at https://nlp.stanford.edu/projects/glove/.}.

\subsection{Post-Processing Word Embeddings}
[9] presents a simple post-processing algorithm that renders off-the-shelf word embeddings even stronger, as measured on a number of lexical-level and sentence-level tasks. The algorithm is based on the geometrical observations that the word embeddings (across all representations such as Glove, word2vec etc.) have a large mean vector and most of their energy (after subtracting the mean vector) is located in a subspace of about 8 dimensions. Since, all embeddings share a common mean vector and all embeddings have the same dominating directions, both of which strongly influence the representations \textbf{in the same way}, eliminating them renders the embeddings stronger. A formal description of the post-processing algorithm from [9] is presented below:

\begin{framed}
  \begin{center}
    \begin{minipage}{0.6\linewidth}
\begin{algorithm}{\normalfont{The Post-Processing Algorithm, PPA(X, D)}}


  \begin{algorithmic}[0]
      \scriptsize
      \State \textbf{Input}: Word Embedding Matrix X, Threshold Parameter D.\\
	  \State \textbf{1. Subtract the Mean}:\\ \hspace{10mm} X = X - mean(X). 

    \State \textbf{2. Compute the PCA Components}: \\ \hspace{10mm} $ u_i$ = PCA(X), where i = 1,2,...,d.

    \State \textbf{3. Eliminate the Top D Components}: $ \forall $ v in X:\\ \hspace{10mm} $ v = v -  \sum_{i=1}^{D}(u_i^T \cdot v)u_i $  \\
           \State \textbf{Output}: Post-Processed Word Embedding Matrix X.
  \end{algorithmic}
\end{algorithm}

    \end{minipage}
  \end{center}

\end{framed}

\textbf{Figure 1(a)} demonstrates the impact of the post-processing algorithm (PPA, with $D=7$) as observed on Glove embeddings (300-dimensions). It compares the fraction of variance explained by the top 20 principal components of the original and post-processed word vectors respectively (the total sum of explained variances over the 300 principal components is equal to 1.0). In the post-processed word embeddings none of the top principal components are disproportionately dominant in terms of explaining the data, which implies that the post-processed word vectors are not as influenced by the common dominant directions as the original embeddings. This makes the individual word vectors more ``discriminative", hence, improving their quality, as validated on several benchmarks in [9].

\subsection{Dimensionality Reduction}
\begin{figure}[!tbp]
  \centering
  \begin{minipage}[b]{\textwidth}
    \includegraphics[width=0.582\textwidth]{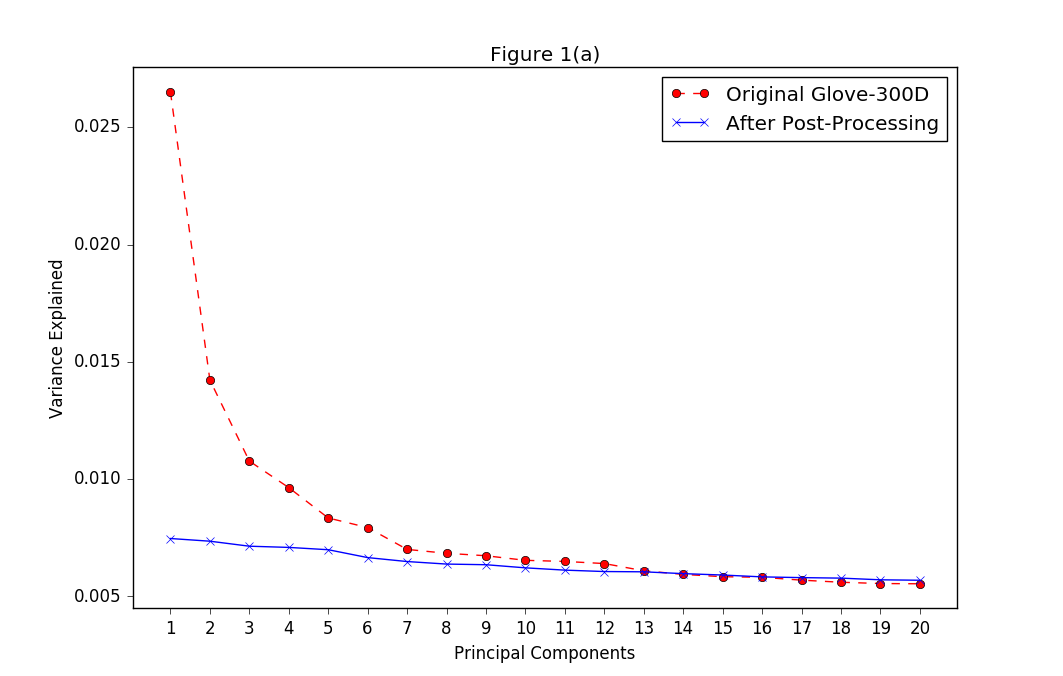}
    \includegraphics[width=0.549\textwidth]{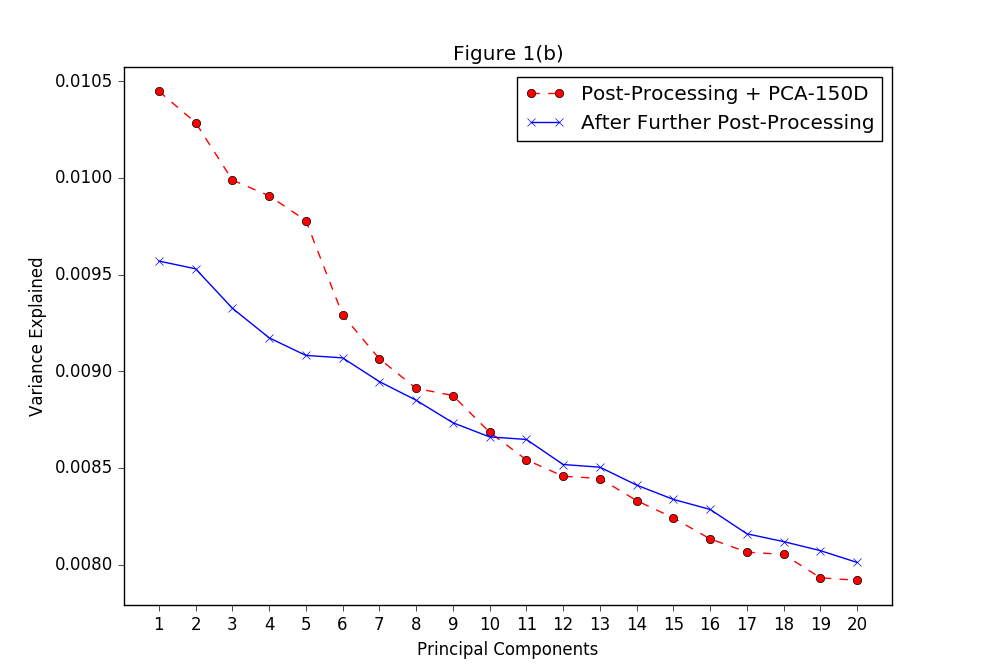}
    \caption{Comparison of (a) the Original and Post-Processed Glove Embeddings (300-Dimensional) in terms of fraction of variance explained by top 20 Principal Components. (b) the Post-Processing + PCA-150D Baseline and Further Post-Processed Glove Embeddings (150-Dimensional) in terms of fraction of variance explained by top 20 Principal Components.}
  \end{minipage}
\end{figure}

In this section we explain and present our algorithm that effectively incorporates the post-processing algorithm in the dimensionality reduction procedure. First of all, since the post-processing algorithm demonstrably leads to better word embeddings, it is appropriate that to construct a lower dimensional representation of word embeddings, the dimensionality reduction algorithm (PCA [13]) be applied on the ``purified" word embeddings.

Secondly, to motivate our algorithm, consider \textbf{Figure 2(b)}. It compares the variance explained by the top 20 principal components for the embeddings constructed by first post-processing the Glove-300D embeddings according to Algorithm 1 (PPA) and then transforming the vectors to 150 dimensions using PCA (labelled as Post-Processing + PCA); against a further post-processed version of the same embeddings (the total sum of explained variances over the 150 principal components is equal to 1.0). We observe that even though PCA has been applied on post-processed embeddings (which had their dominant directions eliminated), the variance in the resulting embeddings is still explained disproportionately by a few of the top principal components. The re-emergence of this geometrical behaviour implies that further post-processing the lower-dimensional embeddings by projecting the word vectors away from the dominant directions will make the embeddings even stronger.

Finally, from Figures 1(a) and 1(b), it is also evident that the extent to which the top principal components explain the data in the case of the reduced embeddings is not as great as in the case of the original 300 dimensional embeddings. Hence, multiple levels of post-processing at different levels of dimensionality will yield diminishing returns as the influence of common dominant directions decrease on the word embeddings.  

\begin{framed}
  \begin{center}
    \begin{minipage}{0.6\linewidth}

\begin{algorithm}{{\normalfont The Dimensionality Reduction Algorithm}}
   

  \begin{algorithmic}[0]
     \scriptsize  
  \State \textbf{Input}: Word Embedding Matrix X, New Dimension N, Threshold Parameter D. \\
  
   \State \textbf{1. Apply the Post-Processing Algorithm}:\\ \hspace{10mm} X = PPA(X, D).
   
    \State \textbf{2. Transform X Using PCA}:\\ \hspace{10mm} X = PCA(X). 
    
    \State \textbf{2. Apply the Post-Processing Algorithm}:\\ \hspace{10mm} X = PPA(X, D). \\ 
    
    \State \textbf{Output}: Word Embedding Matrix of Reduced Dimension N: X.

  \end{algorithmic}
\end{algorithm}

    \end{minipage}
  \end{center}

\end{framed}

These considerations form the intuition behind our algorithm for constructing lower-dimensional word embeddings, where we apply the post-processing algorithm on either side of a PCA based dimensionality reduction of the word vectors. A formal description of our algorithm is presented as Algorithm 2 (further comments on choosing an appropriate value for parameter $D$ are presented in subsection 2.3.4). Further, in our implementation, subtracting the common mean vector from word embeddings is also done as a pre-processing step before applying PCA on the embedding matrix.  

\subsection{Evaluation}
\subsubsection{Word Embeddings}

\begin{table*}
\centering
\caption{Dataset Statistics and Performance (Spearman's Correlation Coefficient x 100) of Various Algorithms on 300 Dimensional Glove Vectors.}
\begin{tabular}{|c|c|c|c|c|c|c|c|c|} \hline
SN &Dataset-Name&WP&Glove-300D&PCA-150D&P+PCA-150D&PCA-150D+P&\textbf{Algo-150D}\\ \hline
1 & MTurk-771 & 771 & 65.01 &  52.47 & \textbf{65.59}&  63.86 & 64.58 \\ \hline
2 & WS-353-SIM & 203 & 66.38 &  52.69 & 70.03 & 70.87  & \textbf{71.61} \\ \hline
3 & MTurk-287 & 287 & 63.32 & 56.56 & 63.38 & \textbf{64.62}  & 63.01 \\ \hline
4 & VERB-143 & 144 & 30.51 & 28.52 & 39.04 & 40.14  & \textbf{42.24} \\ \hline
5 & WS-353-ALL & 353 & 60.54 & 46.52 & 66.23 & 66.85 & \textbf{67.41}  \\ \hline
6 & RW-Stanford & 2034 & 41.18 & 27.46 & \textbf{43.17} & 40.79 & 42.21 \\ \hline
7 & MEN-TR-3K & 3000 & 73.75 & 63.35 & 75.34 & 75.37 & \textbf{75.80} \\ \hline
8 & RG-65 & 65 & \textbf{76.62} &  71.71 & 73.62 & 74.27 & 75.71  \\ \hline
9 & MC-30 & 30 & 70.26 & 70.03 & 69.21 & 72.35 & \textbf{74.80}  \\ \hline
10 & SIMLEX-999 & 999 & \textbf{37.05} &  27.21 & 36.71 & 33.81 & 35.57  \\ \hline
11 & WS-353-REL & 252 & 57.26 & 41.82 & 62.02 & 60.50  & \textbf{62.09}  \\ \hline
12 & YP-130 & 130 & \textbf{56.13} & 36.72 & 55.42 & 50.20 & 55.91 \\
\hline\end{tabular}
\end{table*}

\begin{table*}
\centering
\caption{Performance (Spearman's Correlation Coefficient x 100) of Algorithm 2 across different Embedding Types and Dimensions}
\begin{tabular}{|c|c|c|c|c|c|c|c|c|} \hline
Serial No.&FastText-300D&Algo-150D&Glove-100D&Algo-50D&Glove-200D&Algo-100D\\ \hline
1 & 66.89 & \textbf{67.29} & 58.05 & \textbf{58.85} & 62.12  & 61.99\\ \hline
2 & 78.12 & 77.40 & 60.35 & \textbf{66.27} & 62.91  & \textbf{68.43}\\ \hline
3 & 67.93 & 66.17 & 61.93 & \textbf{64.09} & 61.99  & \textbf{63.55}\\ \hline
4 & 39.73 & 34.24 & 30.23 & \textbf{33.04} & 28.45  & \textbf{36.82}\\ \hline
5 & 73.69 & 73.16 & 52.90 & \textbf{62.05} & 57.42 & \textbf{65.41} \\ \hline
6 & 48.66 & 47.19 & 36.64 & 36.64 & 38.95 & \textbf{39.80}\\ \hline
7 & 76.37 & 76.36 & 68.09 & \textbf{70.93} & 71.01 & \textbf{74.44} \\ \hline
8 & 79.74 & \textbf{80.95} & 69.07 & 64.56 & 71.26 & \textbf{71.53}\\ \hline
9 & 81.23 & \textbf{86.41} & 62.71 & \textbf{68.79} & 66.56 & \textbf{69.83}\\ \hline
10 & 38.03 & 35.47 & 29.75 & 29.13 & 34.03 & \textbf{34.19}\\ \hline
11 & 68.21 & \textbf{69.96} & 49.55 & \textbf{59.55} & 54.48 & \textbf{61.56}\\ \hline
12 & 53.33 & 50.90 & 45.43 & 41.95 & 52.21 & 49.94\\
\hline\end{tabular}
\end{table*}

The pre-trained word embeddings (for English only) used for evaluating our algorithms are: Glove embeddings of dimensions 300, 200 and 100, trained on Wikipedia 2014 and Gigaword 5 corpus (400K vocabulary) [12] and fastText embeddings of 300 dimensions trained on Wikipedia using the Skip-Gram model described in [2] (with 2.5M vocabulary). 

\subsubsection{Datasets}

We use the standard word similarity benchmarks summarized in [5] for evaluating the word vectors [5]. The datasets have word pairs (WP) that have been assigned similarity rating by humans. While evaluating word vectors, the similarity between the words is calculated by the cosine similarity of their vector representations. Then, Spearman's rank correlation coefficient (Rho) between the ranks produced by using the word vectors and the human rankings is calculated. The reported metric in experiments is Rho x 100. Hence, for better word similarity, the evaluation metric will be higher.

\subsubsection{Compared Baselines}
To evaluate the performance of our algorithm, we establish some baselines comprising of alternative schemes of combining the post-processing algorithm along with PCA based dimensionality reduction \footnote{Further, in our experiments, generic non-linear dimensionality-reduction/manifold-learning techniques such as kernel PCA, autoencoders performed worse than the baselines, presumably because they do not exploit the unique geometrical property of word embeddings as discussed in subsection 2.2.}. The baseline algorithms are:
\begin{enumerate}
\item \textbf{PCA}: Transform the word vectors using PCA.
\item \textbf{P+PCA}: Apply the post-processing algorithm and then transform the word vectors using PCA. 
\item \textbf{PCA+P}: Transform word vectors using PCA and then apply the post-processing algorithm. 
\end{enumerate}

These baselines can also be regarded as ablations on our algorithm and can shed light on whether our intuitions in developing the algorithm were correct. In the comparisons ahead, our algorithm is represented as Algo-N (where N is the new dimensionality of the word embeddings). For all evaluations, we use the PCA implementation available in [11].

\subsubsection{Experiments and Evaluation Results}
First we evaluate our algorithm on the same embeddings against the 3 baselines mentioned above and then, we evaluate our algorithm across word embeddings of different dimensions and types. In all the experiments, the threshold parameter $D$ in the PPA algorithm was set to $7$ and the new dimensionality after applying the dimensionality reduction algorithms, $N$ is set to $ d/2 $. 

The appropriate value of parameter $ D $ could be inferred from the "Variance Explained" vs "Principle Components" plot as in Figure 1. It could be observed that the top 7 components are disproportionately contributing to the variance. Hence, setting $ D=7 $ is appropriate as choosing a lower $D$ will not eliminate the disproportionately dominant directions, while choosing a higher D will eliminate useful discriminative information from the word vectors. Further, we show the results at $ N = d/2 $ since at this dimension, the reduced embeddings consistently outperform/achieve very close results to the original embeddings. We observed that going below half the dimensions ($ N < d/2$) significantly hurts the performance for all of the baselines. 

\textbf{Against Different Baselines}: \textbf{Table 1} summarizes the results of different baselines on the 12 datasets. As expected from the discussions in Section 2.1, our algorithm achieves the best results on 6 out of 12 datasets when compared across all the columns (the best scores are highlighted in bold).
In particular, the 150-dimensional word embeddings constructed using our algorithm  performs better than the 300-dimensional embeddings in 7 out of 12 datasets (with an average improvement of 2.74\% across the 12 datasets), does significantly better than PCA, PCA+P baselines and beats P+PCA baseline in 8 out of the 12 tasks.

\textbf{Across Different Embeddings (Both Type \& Dimensionality)}: \textbf{Table 2} summarizes the results of applying our algorithm on 300-dimensional fastText embeddings, 100-dimensional Glove embeddings and 200-dimensional Glove embeddings (the better scores are highlighted in bold). In the case of fastText embeddings, the 150-dimensional word vectors constructed using our algorithm gets better performance on 4 out of 12 datasets when compared to the 300-dimensional embeddings. Overall, the 150-dimensional word vectors have a cumulative score of 765.5 against the 771.93 of the 300-dimensional vectors. Hence, overall its performance is quite similar to the 300-dimensional embeddings (with an average performance decline of 0.53\% across the 12 datasets). In the case of Glove embeddings of 100 and 200 dimensions, our algorithm leads to significant gains (with average performance improvements of 2.6\% and 3\% respectively) over the original embeddings and achieves better performance on 8 and 10 datasets respectively. 

Another interesting observation from \textbf{Table 2} is that the embeddings generated by reducing Glove-200D to 100 dimensions using our algorithm (Algo-100D) significantly outperform the original Glove-100D embeddings (with an average performance improvement of 6\% across all the 12 datasets).    

Hence, empirical results validate that our algorithm is effective in constructing lower dimensional word embeddings, while maintaining similar or (more often) better performance than the higher dimension embeddings.

\section{Conclusions}
Our algorithm considerably reduces the size of word embeddings while maintaining or improving upon their utility and has a very simple implementation. This will allow the use of word embeddings on memory-constrained devices by significantly reducing the memory requirements. In future, an interesting area to explore would be the application of compressed and limited precision representations on top of dimensionality reduction to further reduce the size of the word embeddings. We are also working on automating our algorithm, by deriving an algorithm to choose $D$, $N$ and the levels of post-processing automatically, while optimizing for performance.  


\section*{References}

\small

[1] Martin Andrews. 2016. Compressing Word Embeddings, Springer International Publishing, pages 413–422. 

[2] Piotr Bojanowski, Edouard Grave, Armand Joulin, and Tomas Mikolov. 2017. Enriching word166 vectors with subword information. Transactions of the Association for Computational Linguistics 5:135–146.

[3] Tolga Bolukbasi, Kai-Wei Chang, James Y Zou, Venkatesh Saligrama, and Adam T Kalai. 2016. Man is to computer programmer as woman is to homemaker? debiasing word embeddings. In Advances in Neural Information Processing Systems. pages 4349–4357.

[4] Manaal Faruqui, Jesse Dodge, Sujay K Jauhar, Chris Dyer, Eduard Hovy, and Noah A Smith. 2014. Retrofitting word vectors to semantic lexicons. arXiv preprint arXiv:1411.4166.

[5] Manaal Faruqui and Chris Dyer. 2014. at wordvectors. org. ACL 2014 page 19.

[6] Shaoshi Ling, Yangqiu Song, and Dan Roth. 2016. Word embeddings with limited memory. In The175 54th Annual Meeting of the Association for Computational Linguistics. page 387.

[7] Tomas Mikolov, Ilya Sutskever, Kai Chen, Greg S Corrado, and Jeff Dean. 2013. Distributed representations of words and phrases and their compositionality. In Advances in neural information processing systems. pages 3111–3119.

[8] Nikola Mrkšic, Diarmuid OSéaghdha, Blaise Thomson, Milica Gašic, Lina Rojas-Barahona, Pei-Hao Su, David Vandyke, Tsung-Hsien Wen, and Steve Young. 2016. Counterfitting word vectors to linguistic constraints. In Proceedings of NAACL-HLT. pages 142-148.

[9] Jiaqi Mu, Suma Bhat, and Pramod Viswanath. 2017. All-but-the-top: Simple and effective postprocessing for word representations. arXiv preprint arXiv:1702.01417.

[10] Kim Anh Nguyen, Sabine Schulte Walde, and Ngoc Thang Vu. 2016. Neural based noise-filtering from word embeddings. arXiv preprint arXiv: 160.01874. 

[11] Fabian Pedregosa, Gaël Varoquaux, Alexandre Gramfort, Vincent Michel, Bertrand Thirion, Olivier Grisel, Mathieu Blondel, Peter Prettenhofer, Ron Weiss, Vincent Dubourg, etal. 2011. Scikit-learn: Machine learning in python. Journal of Machine Learning Research 12(Oct):2825–2830.

[12] Jeffrey Pennington, Richard Socher, and Christopher D Manning. 2014. Glove: Global vectors for word representation. In EMNLP. Citeseer.

[13] Jonathon Shlens. 2014. A tutorial on principal component analysis. arXiv preprint arXiv:1404.1100.

\end{document}